\documentclass[10pt]{article}

\usepackage[textheight=9in,
            textwidth=6.5in,
            top=1in,
            headheight=14pt,
            headsep=25pt,
            footskip=30pt]{geometry}

\usepackage[T1]{fontenc}    
\usepackage[utf8]{inputenc} 

\usepackage{amsmath}
\usepackage{amssymb}
\usepackage{amsfonts}       
\usepackage{bbm}            
\usepackage{bbding}         
\usepackage{nicefrac}       

\usepackage{array}
\usepackage{tabularx}
\usepackage{multirow}
\usepackage{makecell}
\usepackage{booktabs}       
\usepackage[table]{xcolor}  
\usepackage{colortbl}
\usepackage{arydshln}       
\usepackage{adjustbox}      

\usepackage{graphicx}
\graphicspath{{media/}}     
\usepackage{subcaption}     
\usepackage[labelfont=bf]{caption} 
\usepackage{wrapfig}        

\usepackage[numbers,sort&compress]{natbib}
\usepackage{parskip}
\usepackage{microtype}      
\usepackage{varwidth}       
\usepackage{multicol}       

\usepackage[many]{tcolorbox}
\tcbuselibrary{skins}       
\tcbuselibrary{breakable}   

\usepackage{hyperref}
\hypersetup{
    colorlinks=true,
    linkcolor=blue,
    filecolor=magenta,      
    urlcolor=cyan,
    citecolor=blue,
    pdftitle={Your Title},
    pdfauthor={Your Name}
}

\usepackage{fancyhdr}
\title{SEReDeEP: Hallucination Detection in Retrieval-Augmented Models via Semantic Entropy and Context-Parameter Fusion}
\author{Lei Wang \\ \textit{Hunan University} \\ \textit{lik639259@hnu.edu.cn}}
\date{\today}

\begin{document}

\maketitle

\begin{abstract}
\textit{Retrieval-Augmented Generation (RAG) models frequently encounter hallucination phenomena when integrating external information with internal parametric knowledge. Empirical studies demonstrate that the disequilibrium between external contextual information and internal parametric knowledge constitutes a primary factor in hallucination generation. Existing hallucination detection methodologies predominantly emphasize either the external or internal mechanism in isolation, thereby overlooking their synergistic effects. The recently proposed ReDeEP framework decouples these dual mechanisms, identifying two critical contributors to hallucinations: excessive reliance on parametric knowledge encoded in feed-forward networks (FFN) and insufficient utilization of external information by attention mechanisms (particularly copy heads). ReDeEP quantitatively assesses these factors to detect hallucinations and dynamically modulates the contributions of FFNs and copy heads to attenuate their occurrence. Nevertheless, ReDeEP and numerous other hallucination detection approaches have been employed at logit-level uncertainty estimation or language-level self-consistency evaluation, inadequately address the semantic dimensions of model responses, resulting in inconsistent hallucination assessments in RAG implementations. Building upon ReDeEP's foundation, this paper introduces SEReDeEP, which enhances computational processes through semantic entropy captured via trained linear probes, thereby achieving hallucination assessments that more accurately reflect ground truth evaluations.}
\end{abstract}

\section{Introduction}
In contemporary language model development, Retrieval-Augmented Generation (RAG) architectures have demonstrated remarkable efficacy in mitigating hallucination phenomena compared to traditional language models. Empirical evaluations reveal that DeepSeek-V3 exhibits a hallucination rate of 29.67\% in factual assessment protocols, which substantially decreases to 24.67\% upon activation of web search capabilities (effectively implementing RAG methodology)\cite{deepseekai}. ERNIE 4.0 Turbo, exemplifying RAG implementation, attains an impressive non-hallucination rate approximating 83\%, with exceptional performance in factual verification and esoteric knowledge domains (contrasting significantly with non-RAG architectures such as kimi-k1.5, which achieve merely 60\% non-hallucination rates under identical evaluation conditions)\cite{Huang}. The hallucination mitigation efficacy of RAG systems derives from their capacity to augment Large Language Model (LLM) responses with pertinent information extracted from external knowledge repositories\cite{survey}.

\begin{figure}
\centering
\includegraphics[width=1.0\textwidth]{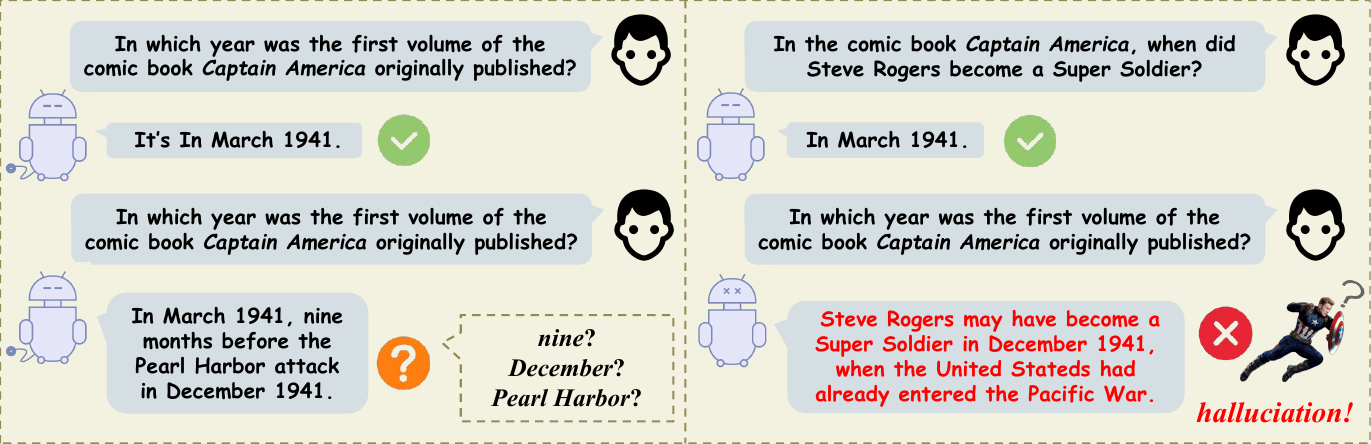}
\caption{\textit{Two examples of language models: The example on the left is a RAG model, where connecting to external knowledge retrieval increases the uncertainty of language expression, intuitively manifesting as outputs with the same semantics but containing some "redundant" vocabulary that interferes with hallucination assessment. The example on the right is a language model without RAG enabled, which produces genuine hallucinations.}}
\setlength{\belowcaptionskip}{-1000pt} 
\label{fig2}
\end{figure}

\begin{figure}
\centering
\includegraphics[width=1.0\textwidth]{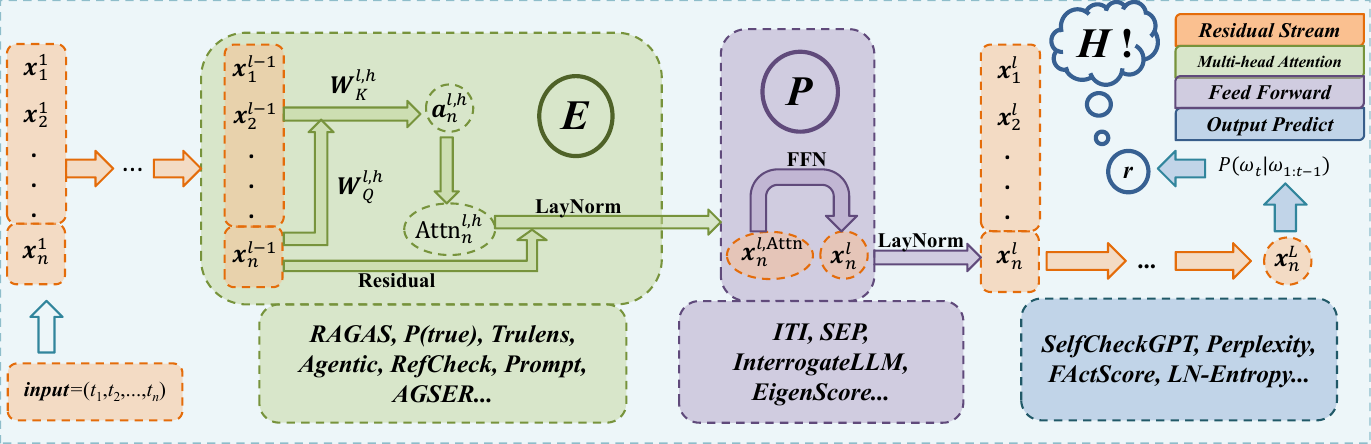}
\caption{\textit{Schematic diagram of residual flow in the decoder. Extracting hallucination features from different parts of the decoder constitutes three mainstream approaches for hallucination detection in current RAG models: treating the attention module as interference factor E, treating the feed-forward network module as interference factor P, and directly sampling at the model's output end. ReDeEP decouples the influences of E and P in a hybrid approach, while INSIDE combines semantic analysis with output-end sampling.}}
\label{fig1}
\end{figure}

Nevertheless, this integration introduces novel challenges, specifically "RAG hallucinations", manifesting as assertions either unsupported by or contradictory to the retrieved informational context\cite{ragtruth}. Initial investigative approaches employed sampling methodologies or uncertainty quantification metrics for hallucination assessment, exemplified by SelfCheckGPT and Perplexity\cite{selfcheckgpt,perplexity}. However, these methodologies remained predominantly phenomenological, failing to elucidate the fundamental etiological mechanisms of hallucinations, while simultaneously incurring substantial computational overhead and error margins\cite{trustmeimwrong}. Adopting mechanistic interpretability\cite{primerinnerworking}, contemporary research increasingly scrutinizes the internal architectural components of models to identify hallucination causal factors, as evidenced in frameworks such as ITI and RAGAS\cite{ITI,RAGAS}. The genesis of model hallucinations constitutes a multifaceted process, with recent investigations indicating antagonistic interactions between external contextual information and internal parametric knowledge within RAG systems\cite{lookWithin,knowledge}. For external context (E) and internal parametric knowledge (P), both elements synergistically influence hallucination (H) manifestation. Currently, multiple causal analytical frameworks facilitate RAG hallucination detection. From the external contextual perspective, these methodologies predominantly analyze decoder attention mechanisms and activation layer hidden representations, establishing causal trajectories from E to H\cite{chyzhyk2022remove,lookWithin}; conversely, from the internal knowledge perspective, certain approaches meticulously examine feed-forward network (FFN) influences within LLMs, either through comparative analysis of residual flow transformations pre- and post-FFN processing, or through input condition modulation to observe intermediate computational state variations, establishing causal trajectories from P to H. The ReDeEP framework further advances this domain by internally decoupling E and P influences, proposing dynamic calibration of their respective contributions to attenuate hallucination phenomena\cite{ReDeEP}.\textbf{Figure} \textbf{\ref{fig1}} shows the locations that these methods focus on. Comprehensive methodological expositions are provided in the Appendix \ref{sec:Baseline}.

From a mechanistic interpretability perspective, while RAG hallucinations manifest in generated output content, they remain detectable within the model's hidden representational states, with detectability progressively increasing through deeper model layers\cite{explainability}. Consequently, both external context E and internal knowledge P can be quantitatively assessed through extraction of activation values, self-attention weight matrices, and residual flow states across various model layers\cite{lookWithin}. ReDeEP implements a multivariate analytical approach, predicting hallucination phenomena through regression analysis of decoupled external context and parametric knowledge metrics, achieving state-of-the-art performance in RAG hallucination prediction\cite{ReDeEP}.

However, these investigative approaches disproportionately emphasize token-level output representations while inadequately addressing semantic dimensions. Due to linguistic grammatical diversity and inter-sentential characteristics, semantically equivalent outputs may exhibit substantially divergent intermediate representational states, resulting in heterogeneous expressions as illustrated in \textbf{Figure \ref{fig2}}\cite{azaria2023internal}. This engenders the paradoxical phenomenon wherein semantically identical outputs receive dramatically different hallucination scores, occasionally diverging by more than 50\%. Regrettably, the aforementioned methodologies have insufficiently addressed this issue, precipitating widespread "semantic deviation" in hallucination detection, wherein veridical responses may be erroneously classified as hallucinations. While linguistic expressional diversity and customization according to user preferences remain desirable objectives—indeed constituting foundational motivations for RAG systems' external knowledge retrieval—the occurrence of "semantic deviation" remains inevitable\cite{SE,ye2023cognitive,kalai2024calibrated}.INSIDE\cite{inside} proposed a sampling-based semantic evaluation method that utilizes the eigenvalues of the response covariance matrix to measure the semantic consistency and diversity of responses in a dense embedding space. However, this requires additional inference costs to generate as many outputs as possible to ensure the accuracy of hallucination detection, resulting in significantly increased computational costs.

The introduction of the semantic entropy\cite{SE} concept provides a solution to this problem. On one hand, the semantic entropy model extends from measuring token-level lexical expression to semantic expression at the output level. On the other hand, semantic entropy probes can directly capture the semantic entropy of hidden states in a single forward propagation, avoiding the high costs of extensive sampling at the expense of minimal precision loss. Using semantic entropy as a hallucination assessment metric essentially eliminates the differences in hallucination scores caused by heterogeneous expressions, greatly improving the precision of hallucination detection to nearly the level of human annotation. Semantic entropy can also enhance the accuracy of conflict hallucination detection in LLMs, and hallucination suppression based on this can further improve the generalization ability of RAG models.

Motivated by semantic entropy theoretical frameworks, this research explores the integration of semantic entropy in concurrent analysis of external context and internal parametric knowledge. By incorporating semantic entropy theory into RAG hallucination detection mechanisms, we extend ReDeEP's conceptual foundation of dynamically decoupling external context and parametric knowledge contributions. Building upon this foundation, we have developed semantic entropy-related detection methodologies to supersede corresponding metrics in the original ReDeEP framework. Experimental validation demonstrates accuracy improvements of 3\%-10\% across relevant datasets compared to the baseline ReDeEP implementation.

\section{Related works: Transformer, EP and SE}
\subsection{Transformer Decoder}
\textbf{Decoder Structure.} Our research primarily focuses on the decoder architecture of language models, which constitutes the fundamental entry point for investigating hallucination generation in retrieval-augmented models\cite{vaswani2017attention}. The decoder in RAG systems represents an integral component of the Transformer architecture, with mainstream language models such as GPT being predicated on transformer decoder structures. Specifically, each decoder layer incorporates a self-attention module (Attn) and a feed-forward network module (FFN). Within these layers, residual\cite{elhage2021mathematical,ferrando2024information,wu2024retrieval} connections integrate information from self-attention heads and feed-forward networks, generating hidden states. These hidden states encapsulate rich semantic information capable of reflecting the intrinsic characteristics of decoder-generated hallucinations\cite{explainability,zou2023representation,subramani2022extracting,hernandez2023inspecting,lookWithin}. In the terminal hidden state, vectors undergo one-to-one mapping to the vocabulary space via a softmax function, establishing a vocabulary probability distribution\cite{vaswani2017attention,zheng2025self}. Finally, search algorithms (such as greedy search or beam search) are employed to predict the definitive sequence, culminating in the decoding process\cite{boulanger2013audio,sutskever2014sequence,graves2012sequence}.

\textbf{Feed-Forward Network Module (FFN).} The FFN module invariably succeeds the self-attention heads. Superficially, it executes conventional linear transformations and activations on the current state to derive enriched feature representations. Fundamentally, FFN fulfills an implicit knowledge storage function, with its parameter space encoding substantial internal knowledge acquired during pre-training\cite{geva2020transformer,survey}. To analyze its operational mechanism, extant research frequently employs comparative analytical methodologies, inferring knowledge activation patterns through observation of vocabulary distribution differentials before and after FFN processing of hidden states\cite{geva2020transformer,sukhbaatar2019augmenting,wang2024wise}.

\subsection{External Context and Internal Knowledge Decoupling}
ReDeEP\cite{ReDeEP} represents a hallucination assessment and mitigation mechanism predicated on the FFN module and self-attention heads within the decoder structure. Related empirical investigations indicate that RAG hallucinations originate from insufficient utilization of external knowledge by copy heads in self-attention mechanisms and excessive dependence on parametric knowledge by FFN modules that store internal knowledge\cite{knowledge}. ReDeEP detects hallucinations in language models through regression-based decoupling of external context and internal parametric knowledge, and suppresses hallucinations by recalibrating the contributions of knowledge FFNs and copy heads to the residual flow (augmenting attention while reducing FFN influence)\cite{ferrando2024primer}. To quantify the impact of external context and internal parametric knowledge, two metrics are proposed: External Context Score (ECS) and Parametric Knowledge Score (PKS)\cite{luo2024large,inside}.

\textbf{External Context Score (ECS).} Select the top k\% tokens with the highest attention scores in the external context as attended tokens:
\begin{equation}
    \mathcal{I}_n^{l,h}=\text{arg}\,\text{top}_{k\%}(a_{n}^{l,h})
    \label{I}
\end{equation}
where $a_{n}^{l,h}$ represents the attention weights of the context, calculated by feeding the hidden states of the current token and context into the QK (query and key) circuit for attention computation, involving weight matrices $\boldsymbol{W}_Q^{l,h}\in\mathbb{R}^{d \times d_h}$ and $\boldsymbol{W}_K^{l,h}\in\mathbb{R}^{d \times d_h}$. The detailed calculation process can be found in equation \ref{eq:a} in the appendix \ref{Sec:A}.

Derive ECS through similarity computation between the most recent token in the final hidden layer and historical context tokens. More specifically, we perform mean pooling on the hidden states of the context, then calculate the cosine similarity between this pooled representation and the hidden state of the most recent token, which constitutes the ECS score for that token. This process is applied to each new token.
\begin{gather}
    \mathcal{E}_n^{l,h} = \frac{\boldsymbol{e} \cdot \boldsymbol{x}_n^l}{\|\boldsymbol{e}\| \|\boldsymbol{x}_n^l\|}, \quad \boldsymbol{e} = \frac{1}{|\mathcal{I}_n^{l,h}|} \sum_{j \in \mathcal{I}_n^{l,h}} \boldsymbol{x}_j^l 
    \label{ECS}
\end{gather}

\textbf{Parametric Knowledge Score (PKS).} Let $\boldsymbol{x}_n^{l,\text{Attn}}$ represent the hidden state of the most recent token's residual flow after passing through the attention layer at layer $l$, but before being injected into the FFN. Utilize LogitLens\cite{wang2024wise} to map the residual flow states before and after the FFN layer back to vocabulary distributions, compare their disparities, and measure using Jensen-Shannon divergence (JSD):
\begin{gather}
    \mathcal{P}_n^{l}=\text{JSD}(\text{softmax}(\text{LogitLens}(\boldsymbol{x}_n^{l,\text{Attn}})\,||\,\text{softmax}(\text{LogitLens}(\boldsymbol{x}_n^{l})))
    \label{PKS}
\end{gather}

Regression Decoupling. Integrate ECS and PKS as dual covariates and employ linear regression to predict ultimate hallucination\cite{kahlert2017control}:
\begin{equation}
    \mathcal{R}(\mathbf{r}) = \frac{1}N \sum_{j=1:N} \mathcal{R}(\boldsymbol{x}_j), \quad \mathcal{R}(\boldsymbol{x}_n) = \sum_{l \in \mathcal{F}} \alpha \cdot \mathcal{P}_n^l - \sum_{l,h \in \mathcal{A}} \beta \cdot \mathcal{E}_n^{l,h}
\end{equation}
where $\alpha$ and $\beta$ represent regression coefficients, $\mathcal{F}$ and $\mathcal{A}$ denote specific layers exerting significant influence on the utilization of external context and internal parametric knowledge.This is the token-level working principle of ReDeEP. Chunk-level ReDeEP is used to ensure efficiency when processing slightly longer text content, which is provided in the appendix \ref{Sec:ReDeEP}.

\subsection{Semantic Entropy\cite{SE}}
As mentioned in the introduction, most methods, including ReDeEP, only assess hallucinations at the token level while ignoring differences in expressions\cite{SE,ye2023cognitive,kalai2024calibrated,inside}. Fortunately, the introduction of semantic entropy and semantic entropy probes provides\cite{SEP} us with an optimization approach. Research on semantic entropy probes reveals that the hidden states of language models can still effectively capture semantic entropy information of generated content. In this case, semantic entropy probes can be well embedded into the hallucination assessment criteria proposed by ReDeEP, providing a more scientific, precise, and cost-effective solution for hallucination assessment and suppression in RAG.

\textbf{Semantic Entropy of Large Language Models.} Before semantic entropy was proposed, traditional model hallucination assessment methods often focused only on token-level lexical semantics (consistent with the predictive nature of decoder structures), while ignoring differences in expressions with the same semantics. Sometimes, despite having the same semantics in responses, they might be viewed as different answers due to differences in expression patterns (often brought about by grammatical differences). The semantic entropy model commands language models to generate answers multiple times, forming semantically equivalent clusters through bidirectional entailment relationships. The semantic entropy model enables the assessment of model hallucinations without relying on prior knowledge, abandoning the deep learning paradigm in favor of classical probabilistic machine learning methods and clustering ideas to solve problems. Typically, we use discrete semantic entropy, a variant of semantic entropy, which only calculates the frequency of each semantically equivalent cluster as its probability distribution, used to estimate semantic entropy under unsupervised conditions.
The calculation process of semantic entropy requires temperature sampling and bidirectional entailment relationship confirmation. For a given input context $x$, multiple output sequences and bidirectional entailment clusters are generated through repeated sampling. For each output sequence $s=(t_1,t_2,...t_n)$, the product of conditional token probabilities constitutes the joint probability of the output tokens. The sum of probabilities of all possible generations in the corresponding cluster forms the semantic cluster probability. The uncertainty of the semantic cluster probability distribution is the semantic entropy:
\begin{gather}
    \text{SE}({x}) = -\sum_{C} p(C \,|\, x) \log p(C\,|\, x)\\    
    p(C\,|\,x)=\sum_{s \in C}p(s\,|\,x)\\
    p(s\,|\,x)=\prod_{i=1}^{n}p(t_i\,|\,t_{1:i-1},x) \label{SE1}
\end{gather}
Considering computational cost, discrete semantic entropy is typically adopted in practical applications, establishing a semantic clustering distribution model $\mathcal{C}=(C_1,C_2,...,C_k)$ based on Monte Carlo sampling, approximating semantic entropy as:
\begin{equation}
    \widetilde{SE}(x) \approx -\frac{1}{|C|} \sum_{{C}_m \in C} p(C_m \,|\, x)\log p(C_m \,|\, x) \label{SE2}
\end{equation}

\textbf{Probes: Cost-effective and Reliable Semantic Entropy Capture Method.\cite{SEP}} Semantic entropy probes were proposed to address the high computational cost of semantic entropy calculation. As mentioned earlier, the semantic entropy model only evaluates based on generated content at the output end of language models, without considering deep learning patterns. In contrast, semantic entropy probes refocus on the representation of hidden states in language models, training linear probes on hidden states obtained from a single forward pass to directly approximate semantic entropy, without requiring sampling of multiple responses at the model's output end\cite{semantic}.
The inspiration for the probe idea comes from the use of splitting objectives in regression trees\cite{breiman2017classification}, binarizing semantic entropy scores into labels representing high or low semantic entropy, and training logistic regression classifiers to predict these labels, using the probability of high values returned by the logistic regression classifier as the predicted semantic entropy. This probe reduces computational complexity by two orders of magnitude while maintaining over 90\% evaluation accuracy\cite{SEP}.

More detailed principles are provided in the appendix \ref{Sec:SEP}.In summary, we can approximate semantic entropy using the prediction results from probes on hidden layers. Research indicates that after the 9th hidden layer, semantic entropy probes can rapidly capture semantic entropy while sacrificing only about 10\% of precision\cite{SEP,semantic}.More importantly, we can observe the global changes of semantic entropy across the entire decoder, rather than just at the output end as described in equations \ref{SE1} and \ref{SE2}.

\section{Method}
Based on the aforementioned empirical principles and related experimental findings, we propose SEReDeEP, a framework that preserves ReDeEP's fundamental approach of decoupling external context and internal parametric knowledge while introducing semantic entropy theory and semantic entropy probe tools to reconstruct the computational processes for external context scores and internal parametric knowledge scores. It is important to note that while our computational methodology is applicable across all hidden layers in RAG architectures, we preferentially conduct calculations on hidden layers subsequent to the 9th layer. This preference is justified by several factors: compared to layers 1-9, deeper hidden layers manifest richer semantic representations\cite{nguyen2025overcoming,sun2024transformer,radford2018improving}; the probabilities predicted by probes at these deeper layers approximate true semantic entropy with greater fidelity\cite{SEP}; and this approach ensures consideration of sufficient hidden layers to prevent loss of critical information from the external context\cite{sun2024transformer,vaswani2017attention}.

\textbf{External Context Entropy Score(ECE).} For the current output token $t_n$ with hidden state representation $x_n^l$, we employ specialized probes to capture the semantic entropy of this hidden state:
\begin{equation}
\widetilde{H}_n^{l}=\widetilde{SEP}(\boldsymbol{x}_n^l)
\end{equation}
where $\widetilde{SEP}$ denotes our semantic entropy probe, with detailed training methodology elaborated in the appendix \ref{Sec:SEP}.

For attended tokens in the context (as derived from equation \ref{I}), we independently detect semantic entropy values:
\begin{gather}
    \widetilde{H}_{\leq n-1}^ {l,h}=(\widetilde{SEP}(\boldsymbol{x}_j^l)\,|\,j\in \mathcal{I}_n^{l,h}) 
\end{gather}

We quantify the semantic correlation between the current state and the context by computing Z-Score between the current hidden state's semantic entropy and context semantic entropy value, aggregating these coefficients to derive the external context score:
\begin{gather}
    \mathcal{E}_n^{l} =\frac{\widetilde{H}_n^l-\mu}{\sigma}
    \label{ECE}
\end{gather}
where $\mu$ is the mean of $\widetilde{H}_n^l$,and $\sigma$ is
the standard deviation of $\widetilde{H}_n^l$.
Our approach was inspired by the use of Z-Score to measure the intensity of information interaction between different attention mechanisms and the correlation between attention heads\cite{voita2019analyzing}.

\textbf{Parametric Knowledge Entropy Score(PKE).} Feed-forward networks (FFNs) function as repositories of internal parametric knowledge\cite{geva2020transformer,survey}. We independently predict the semantic entropy corresponding to hidden states before and after FFN layer processing to derive the internal parametric knowledge score:
\begin{equation}
\mathcal{P}_n^{l}=\lvert \widetilde{SEP}(\boldsymbol{x}_n^{l,\text{Attn}})-\widetilde{SEP}(\boldsymbol{x}_n^{l}) \lvert
\label{PKE}
\end{equation}
This methodological approach is justified by dual considerations. First, hallucinations arising from excessive reliance on internal parameters frequently manifest semantic contradictions and ambiguities, resulting in either abnormally concentrated semantics (low entropy) or contextual deviation (high entropy) in generated content\cite{press2021train}. Second, compared to ReDeEP's approach of directly utilizing differences in mapped vocabulary distributions as indicators\cite{ReDeEP}, our methodology eliminates dependence on vocabulary coverage capabilities, enabling more effective differentiation between legitimate knowledge and hallucinations. The empirical validation of this parametric knowledge score formulation is substantiated through comprehensive experiments detailed in the appendix.

\textbf{Semantic Entropy Hallucination Score.} We maintain the regression decoupling paradigm to predict the definitive hallucination score\cite{kahlert2017control}:
\begin{equation}
\mathcal{R}(\boldsymbol{r}) = \sum_{l \in \mathcal{F}} \alpha \cdot \mathcal{P}_n^l - \sum_{l,h \in \mathcal{A}} \beta \cdot \mathcal{E}_n^{l,h}
\end{equation}
where $\alpha$ and $\beta$ represent regression coefficients optimized for the specific model architecture and dataset characteristics.

\begin{figure}
\centering
\includegraphics[width=1.0\textwidth]{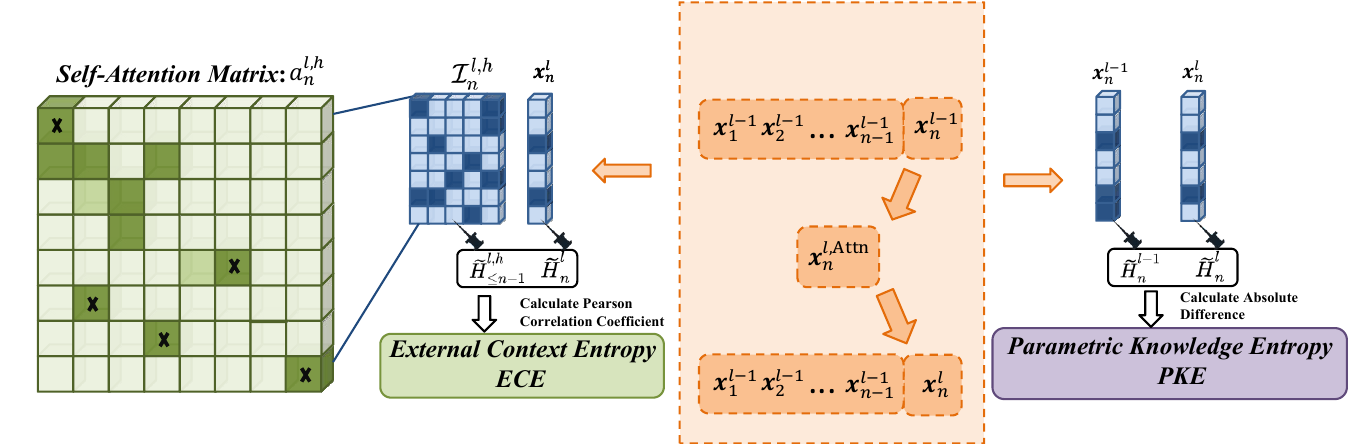}
\caption{\textit{Schematic diagram of SEReDeEP's working principle, with the decoder's residual flow direction in the center, the calculation process of External Context Entropy (ECE) at the left side, and the calculation process of Parametric Knowledge Entropy (PKE) at the right side.}} 
\label{fig3}
\end{figure}

\section{Experience}
\subsection{Experimental Setup}
\textbf{Models.} We conducted comprehensive experiments across three state-of-the-art language models: LLaMA3-7b\cite{metallama2024}, Mistral-7B\cite{mistral2024}, and Qwen2.5-7B\cite{Qwen2024}, utilizing these architectures to generate both hidden representational states and textual responses. For semantic entropy calculation and entailment relationship analysis, we employed DeepSeek-R1\cite{DeepSeek2025} and Claude-3.7-Sonnet-Thinking\cite{Claude2025}, selected specifically for their capacity to generate outputs with accompanying detailed reasoning processes, facilitating transparent observation of computational accuracy throughout our experimental procedures.

\textbf{Datasets.} Our experimental framework incorporated multiple specialized RAG hallucination datasets, including: (1) RAGTruth\cite{wu2023ragtruth}—a meticulously curated corpus designed explicitly for analyzing and evaluating word-level hallucinations in large language models operating within retrieval-augmented frameworks; and (2) HalluRAG\cite{ridder2024hallurag}—a dataset focused on closed-domain hallucination detection in RAG systems, providing granular sentence-level hallucination annotations. Beyond their primary evaluation function.Neither RAGtruth nor HalluRAG contains generated responses from the Qwen2.5-7B model. We randomly extracted questions and retrieved documents from the original datasets, input them into the Qwen model to generate answers, and conducted hallucination assessment through manual annotation.These datasets' original outputs served as critical training material for our semantic entropy probes\cite{SEP}.

\textbf{Semantic Entropy Probes.} The probe training corpus encompassed output responses from both the RAGTruth and HalluRAG datasets, supplemented with semantic entropy measurements provided by DeepSeek-R1 and Claude 3.7 Sonnet Thinking. We wrote a script to batch obtain semantic entropy through pre-defined prompts. Following successful training, these probes were strategically embedded within the decoder layers of the Transformer architectures across all experimental models (LLaMA3-7b\cite{metallama2024}, Mistral-7B\cite{mistral2024}, and Qwen2.5-7B\cite{Qwen2024}).More detailed process are provided in the appendix \ref{Sec:SEP}.

\textbf{Baseline Methods.} To establish comprehensive comparative benchmarks, we implemented ten advanced hallucination detection methodologies, including AGSR(Attention-guided Self-reflection)\cite{AGSR}, Ragas\cite{RAGAS}, ITI\cite{ITI}, INSIDE\cite{inside}, InterrogateLLM\cite{yehuda2024interrogatellm}, FActScore\cite{min2023factscore}, and SelfCheckGPT\cite{selfcheckgpt}, alongside the foundational ReDeEP\cite{ReDeEP} and SEP\cite{SEP} approaches. Detailed specifications and implementation particulars for each baseline method are documented in the appendix \ref{sec:Baseline}.

\subsection{Implementation Details.} 
Our implementation predominantly adhered to default model parameterizations. We leveraged LangChain\cite{langchain} for constructing model-specific RAG agents, employed ChromaDB\cite{chromadb_pypi} as the vector database infrastructure for information retrieval operations, and integrated document processing, vector retrieval, and generation components to implement a unified control flow architecture combining adaptive RAG, corrective RAG, and self-RAG methodologies.

In our experimental model configurations, LLaMA3-7B\cite{metallama2024} and Mistral-7B\cite{mistral2024} featured 32 hidden layers and 32 attention heads, with an architectural design where 4 query heads shared a single key-value head. Qwen2.5-7B\cite{Qwen2024} implemented 28 hidden layers and 28 attention heads, with 7 query heads sharing each key-value head. Consistent with our methodological framework described previously, all experimental measurements were conducted on hidden layers subsequent to the 9th layer.

For attention weight matrix embedding vector selection, we maintained a consistent ratio parameter of $k=10$ across all experimental configurations\ref{I}. Model-specific parameterizations were as follows:

\begin{itemize}
    \item \textbf{LLaMA3-7B\cite{metallama2024}}: For the RAGTruth dataset, we selected the highest-scoring copy head and top 2 FFN layers, with regression coefficients $\alpha=1$ and $\beta=0.2$; for the HalluRAG dataset, we utilized the highest-scoring copy head and top 10 FFN layers, with $\alpha=1$ and $\beta=0.4$.
    
    \item \textbf{Mistral-7B\cite{mistral2024}}: For the RAGTruth dataset, we employed the top 3 copy heads and top 2 FFN layers, with $\alpha=1$ and $\beta=1$; for the HalluRAG dataset, we selected the top 5 copy heads and top 15 FFN layers, with $\alpha=1$ and $\beta=1.6$.
    
    \item \textbf{Qwen2.5-7B\cite{Qwen2024}}: For the RAGTruth dataset, we utilized the highest-scoring copy head and top 7 FFN layers, with $\alpha=1$ and $\beta=0.8$; for the HalluRAG dataset, we selected the top 10 copy heads and top 10 FFN layers, with $\alpha=1$ and $\beta=1.2$.
\end{itemize}

Throughout all experimental configurations, we consistently employed beam search with a length penalty coefficient of 0.8 for response sequence generation\ref{r}.

\subsection{Experimental Results Analysis and Comparison.} 
As evidenced in \textbf{Table \ref{tab:results}}, SEReDeEP demonstrated statistically significant performance improvements across both the RAGTruth and HalluRAG datasets when compared to alternative hallucination detection methodologies, particularly relative to the original ReDeEP and SEP approaches. While certain metrics on specific model configurations exhibited marginally lower performance compared to some baseline methods, these differences were negligible and statistically insignificant.

A notable strength of SEReDeEP lies in its consistent performance across diverse RAG model architectures, demonstrating robust generalization capabilities and operational stability. Quantitatively, our method achieved hallucination prediction accuracy improvements exceeding 3\% compared to ReDeEP and surpassing 10\% relative to SEP across both experimental datasets, underscoring the efficacy of our semantic entropy-based approach to hallucination detection in retrieval-augmented generative systems.

Comparative experiments between SEReDeEP and other methods, with bold indicating the best performance, and underline indicating the second-best performance.
\begin{table}[htbp]
\centering
\fontfamily{ptm}\selectfont
\caption{\centering \textit{Comparative experiments between SEReDeEP and other methods, with bold indicating the best performance and underline indicating the second-best performance.}}
\label{tab:results}
\scriptsize
\setlength{\tabcolsep}{4pt}
\renewcommand{\arraystretch}{1.0}
\begin{tabular}{@{}
>{\centering\arraybackslash}p{8em}@{\hspace{0.5em}}
>{\centering\arraybackslash}p{8em}
*{8}{c}
@{}}
\toprule
\multirow{2.5}{*}{\textbf{Model}} & 
\multirow{2.5}{*}{\textbf{Method}} & 
\multicolumn{4}{c}{\textbf{RAGTruth}\cite{ragtruth}} & 
\multicolumn{4}{c}{\textbf{HalluRAG}\cite{ridder2024hallurag}} \\
\cmidrule(r{3pt}){3-6} \cmidrule(r{3pt}){7-10}
 & & \textbf{ACC} & \textbf{AUC} & \textbf{F1} & \textbf{Rec} & \textbf{ACC} & \textbf{AUC} & \textbf{F1} & \textbf{Rec} \\
\midrule

\multirow{10}{*}{\centering \textbf{LLaMA3-7B}\cite{metallama2024} }
& AGSER\cite{AGSR} & 0.7064 & 0.7112 & 0.6542 & 0.6055 & 0.7054 & 0.6293 & 0.6127 & 0.6102 \\
& RAGAS\cite{RAGAS} & 0.7003 & 0.6928 & 0.6327 & 0.6037 & 0.7983 & 0.7045 & 0.6249 & 0.6122 \\
& ITI\cite{ITI} & 0.6932 & 0.7423 & 0.5824 & 0.6832 & 0.7341 & 0.6231 & 0.5371 & 0.6827 \\
& INSIDE\cite{inside} & 0.7908 & - & \underline{0.7012} & 0.7007 & 0.8123 & - & \underline{0.7321} & 0.6944 \\
& InterogateLLM\cite{yehuda2024interrogatellm} & 0.8057 & 0.7649 & 0.5433 & 0.6039 & 0.8324 & 0.7291 & 0.5823 & 0.6547 \\
& FActScore\cite{min2023factscore} & \underline{0.8553} & 0.7043 & 0.6927 & 0.5829 & 0.8404 & 0.6534 & 0.6621 & 0.6395 \\
& SelfCheckGPT\cite{selfcheckgpt} & 0.7841 & - & 0.3907 & 0.4272 & 0.7556 & 0.6804 & 0.3510 & 0.4928 \\
& SEP\cite{SEP} & 0.7834 & 0.7520 & 0.6674 & 0.7022 & 0.7994 & 0.7433 & 0.6858 & \underline{0.6981} \\
& ReDeEP\cite{ReDeEP} & 0.8323 & $\boldsymbol{0.7829}$ & 0.6910 & $\underline{0.7122}$ & $\boldsymbol{0.8596}$ & $\underline{0.7623}$ & 0.7036 & 0.6905 \\
&  \textbf{SEReDeEP} & $\boldsymbol{0.8601}$ & $\underline{0.7824}$ & $\boldsymbol{0.7237}$ & $\boldsymbol{0.7159}$ & $\underline{0.8478}$ & $\boldsymbol{0.7631}$ & $\boldsymbol{0.7339}$ & $\boldsymbol{0.7162}$ \\
\cmidrule(r{3pt}){1-10}

\multirow{10}{*}{\centering \textbf{Mistral-7B}\cite{mistral2024} }
& AGSER\cite{AGSR} & 0.7288 & 0.7046 & 0.6128 & 0.5326 & 0.7675 & 0.6898 & 0.6555 & 0.4387 \\
& RAGAS\cite{RAGAS} & 0.8043 & 0.6653 & 0.6011 & 0.5932 & 0.7903 & 0.6731 & 0.5321 & 0.5542 \\
& ITI\cite{ITI} & 0.7877 & 0.6844 & 0.5612 & 0.6753 & 0.7131 & 0.5932 & 0.5886 & 0.6941 \\
& INSIDE\cite{inside} & 0.7991 & - & 0.6925 & 0.6742 & 0.8129 & - & \underline{0.7124} & \underline{0.7126} \\
& InterogateLLM\cite{yehuda2024interrogatellm} & 0.7012 & 0.6523 & 0.5672 & 0.5938 & 0.6332 & 0.6531 & 0.5999 & 0.5450 \\
& FActScore\cite{min2023factscore} & 0.7571 & 0.7321 & 0.5816 & 0.6793 & 0.8531 & 0.6129 & 0.5839 & 0.5421 \\
& SelfCheckGPT\cite{selfcheckgpt} & 0.6963 & - & 0.4012 & 0.5013 & 0.7952 & 0.5893 & 0.4311 & 0.4481 \\
& SEP\cite{SEP} & 0.6664 & 0.6927 & \underline{0.6926} & 0.6745 & 0.7874 & 0.6038 & 0.7031 & 0.7036 \\
& ReDeEP\cite{ReDeEP} & \underline{0.8057} & \underline{0.6991} & 0.6511 & \underline{0.6938} & \underline{0.8457} & \underline{0.7074} & $\boldsymbol{0.7234}$ & 0.7054 \\
&  \textbf{SEReDeEP} & $\boldsymbol{0.8257}$ & $\boldsymbol{0.7034}$ & $\boldsymbol{0.6978}$ & $\boldsymbol{0.7169}$ & $\boldsymbol{0.8697}$ & $\boldsymbol{0.7323}$ & 0.7097 & $\boldsymbol{0.7358}$ \\
\cmidrule(r{3pt}){1-10}

\multirow{10}{*}{\centering \textbf{Qwen2.5-7B}\cite{Qwen2024} }
& AGSER\cite{AGSR} & 0.7671 & 0.7329 & 0.5992 & 0.5976 & 0.7532 & 0.6643 & 0.5761 & 0.4869 \\
& RAGAS\cite{RAGAS} & 0.7343 & 0.7042 & 0.6329 & 0.6271 & 0.7845 & 0.6992 & 0.6475 & 0.6071 \\
& ITI\cite{ITI} & 0.7557 & 0.7101 & 0.5120 & 0.6654 & 0.6944 & 0.7019 & 0.5438 & 0.6403 \\
& INSIDE\cite{inside} & \underline{0.8003} & - & 0.6897 & 0.7349 & 0.7967 & - & $\boldsymbol{0.7214}$ & \underline{0.7031} \\
& InterogateLLM\cite{yehuda2024interrogatellm} & 0.7744 & 0.6544 & 0.5231 & 0.6296 & 0.7976 & 0.6111 & 0.5918 & 0.6781 \\ 
& FActScore\cite{min2023factscore} & 0.7062 & 0.6998 & 0.6005 & 0.6638 & \underline{0.8563} & \underline{0.7021} & 0.5912 & 0.6193 \\
& SelfCheckGPT\cite{selfcheckgpt} & 0.7657 & - & 0.3741 & 0.4738 & 0.7940 & 0.6766 & 0.3258 & 0.5024 \\
& SEP\cite{SEP} & 0.7349 & 0.7431 & 0.6828 & 0.6438 & 0.7768 & 0.6080 & 0.6339 & 0.6657 \\
& ReDeEP\cite{ReDeEP} & 0.7984 & \underline{0.7607} & $\boldsymbol{0.7013}$ & \underline{0.6663} & 0.8325 & 0.6932 & 0.7157 & 0.7030 \\
& \textbf{SEReDeEP} & $\boldsymbol{0.8135}$ & $\boldsymbol{0.7839}$ & \underline{0.6951} & $\boldsymbol{0.6908}$ & $\boldsymbol{0.8598}$ & $\boldsymbol{0.7124}$ & \underline{0.7182} & $\boldsymbol{0.7363}$ \\
\bottomrule
\end{tabular}
\end{table}

\clearpage

\subsection{Does Introducing Semantic Entropy Really Better Quantify the Influence of Context and Parametric Knowledge?}
\begin{wrapfigure}{r}{0.66\textwidth}
    \centering
    \setlength{\abovecaptionskip}{3pt}
    \setlength{\belowcaptionskip}{0pt}
    \begin{subfigure}[b]{0.49\linewidth}
        \includegraphics[trim=0 0 0 0, clip, width=\linewidth]{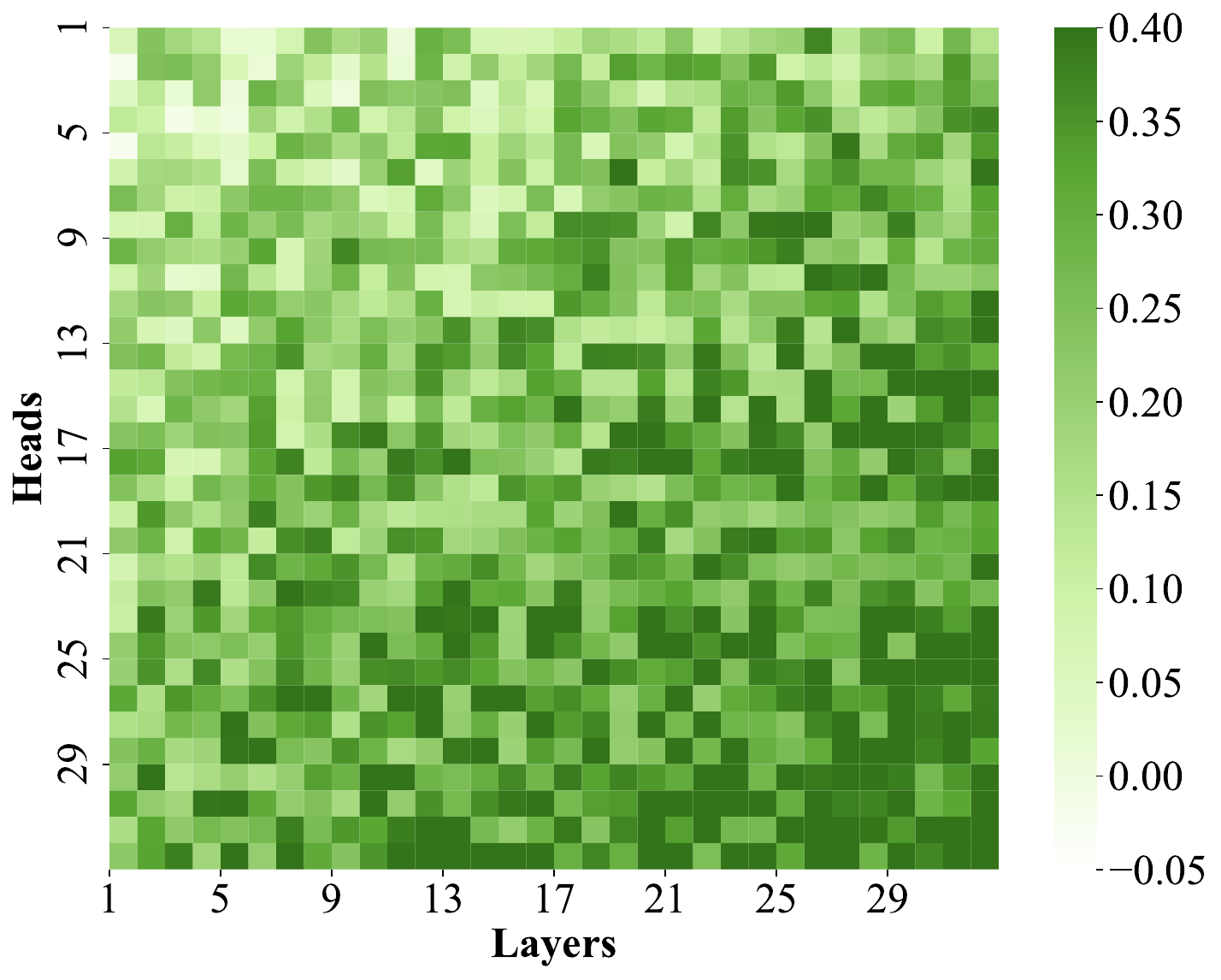}
        \caption{}
    \end{subfigure}
    \hfill
    \begin{subfigure}[b]{0.47\linewidth}
        \includegraphics[trim=0 0 0 0, clip, width=\linewidth]{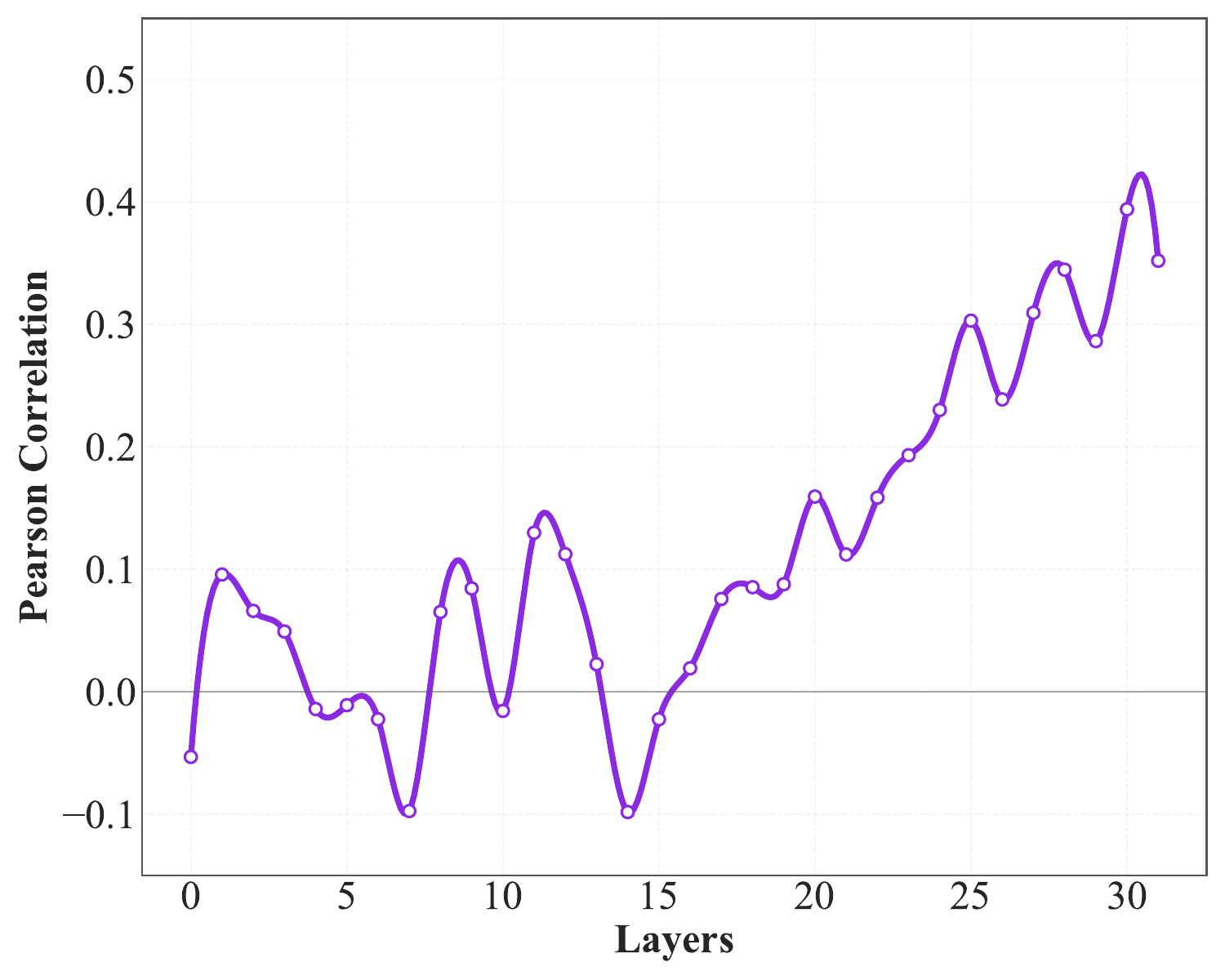}
        \caption{}
    \end{subfigure}
    \caption{\textit{Correlation experimental results between SEReDeEP and hallucinations. (a) shows the correlation coefficient graph between ECE scores of attention heads at each layer and hallucination labels, (b) presents the correlation coefficient graph between PKE scores of feed-forward networks at each layer and hallucination labels.}}
    \vspace{0cm} 
\end{wrapfigure}

Experiments were carried out on LLaMA3-7B\cite{metallama2024}, comparing the correlation between hallucination labels and SEReDeEP's halluciation scores in the RAG-Truth dataset. Specifically, we obtained ECE\ref{ECE} scores(inverted, considering the expected negative correlation) and PKE\cite{PKE} scores from each layer and hallucination labels , calculated the correlation coefficient between the two.

The ECE\ref{ECE} scores of the vast majority of attention heads show a negative correlation with hallucination labels.We observed that the correlation between ECE\ref{ECE} and hallucinations becomes more evident as the number of model layers increases, which is because the precision of the probes gradually improves as the number of model layers increases. The PKE\ref{PKE} scores of most hidden layers exhibit a positive correlation with hallucination labels. This indicates that our semantic entropy probes indeed play an effective role in evaluating external context and parametric knowledge.

\subsection{Ablation}
\label{ablation}
In this section, we will observe the performance of each acting alone, as shown in the figure. The performance of using only ECE\ref{ECE} or only PKE\ref{PKE} is not as good as combining both, and is even worse than most models in \textbf{Table \ref{tab:ablation}}, which is what we expected to see.

Regarding the improvement of ECE\ref{ECE} and PKE\ref{PKE} compared to the original model, we designed more detailed ablation studies in the appendix \ref{More}. The experimental results show that the introduction of semantic entropy improved accuracy in quantifying both external context and parametric knowledge.

\begin{table}[htbp]
\centering
\fontfamily{ptm}\selectfont
\caption{\textit{Ablation of PKE and ECE}}
\label{tab:ablation}
\footnotesize 
\setlength{\tabcolsep}{3pt}
\renewcommand{\arraystretch}{1.1}
\begin{tabular}{@{} 
    >{\centering\arraybackslash}p{8em}
    >{\centering\arraybackslash}c
    >{\centering\arraybackslash}c 
    *{4}{c} 
    *{4}{c} 
@{}}
\toprule[0.8pt]
\multirow{2.5}{*}{\textbf{Model}} & 
\multirow{2.5}{*}{\textbf{PKE} } & 
\multirow{2.5}{*}{\textbf{ECE} } & 
\multicolumn{4}{c}{\textbf{RAGTruth}\cite{ragtruth}} & 
\multicolumn{4}{c}{\textbf{HalluRAG}\cite{ridder2024hallurag}} \\
\cmidrule(lr){4-7} \cmidrule(l){8-11}
&&& 
\textbf{ACC} & \textbf{AUC} & \textbf{F1} & \textbf{Rec} & \textbf{ACC} & \textbf{AUC} & \textbf{F1} & \textbf{Rec} \\ 
\midrule[0.6pt]

\multirow{3}{*}{\textbf{LLaMA3-7B}\cite{metallama2024}} 
& \Checkmark & \XSolidBrush & 0.8403 & 0.7024 & 0.6853 & 0.6955 & 0.8033 & 0.6423 & 0.6945 & 0.6990 \\
& \XSolidBrush & \Checkmark & 0.8592 & 0.7568 & 0.7126 & 0.6735 & 0.8217 & 0.7231 & 0.7123 & 0.7148 \\
& \Checkmark & \Checkmark & 0.8601 & 0.7824 & 0.7237 & 0.7159 & 0.8478 & 0.7631 & 0.7399 & 0.7162 \\
\specialrule{0.4pt}{4pt}{4pt}

\multirow{3}{*}{\textbf{Mistral-7B}\cite{mistral2024}} 
& \Checkmark & \XSolidBrush & 0.8006 & 0.6419 & 0.6792 & 0.6940 & 0.7833 & 0.7029 & 0.6432 & 0.6900 \\
& \XSolidBrush & \Checkmark & 0.7458 & 0.6830 & 0.6571 & 0.7095 & 0.8537 & 0.7200 & 0.6847 & 0.7213 \\
& \Checkmark & \Checkmark & 0.8257 & 0.7034 & 0.6978 & 0.7169 & 0.8697 & 0.7323 & 0.7097 & 0.7358 \\
\specialrule{0.4pt}{4pt}{4pt} 

\multirow{3}{*}{\textbf{Qwen2.5-7B}\cite{Qwen2024}} 
& \Checkmark & \XSolidBrush & 0.6914 & 0.7521 & 0.6543 & 0.6262 & 0.8326 & 0.5834 & 0.6757 & 0.7056 \\
& \XSolidBrush & \Checkmark & 0.7547 & 0.7256 & 0.6812 & 0.6739 & 0.8004 & 0.6782 & 0.6922 & 0.7185 \\
& \Checkmark & \Checkmark & 0.8135 & 0.7839 & 0.6951 & 0.6908 & 0.8598 & 0.7124 & 0.7182 & 0.7363 \\

\bottomrule[0.8pt]
\end{tabular}
\end{table}

\subsection{Intervention Experiments}
We conducted intervention experiments to demonstrate that ECE and PKE indeed quantify the model's utilization of external context and internal knowledge, rather than evaluating hallucinations through spurious correlations. The intervention experiments were implemented by comparing the prediction results of SEDeDeEP and ReDeEP before and after interventions. For external context, we adopted noise intervention, injecting Gaussian noise into the attention weight matrix to add instability to the residual flow; for internal knowledge, we set the activation values of 8 randomly sampled FFNs to zero, thereby erasing their contributions to the residual flow.
\begin{wrapfigure}[12]{r}{0.33\textwidth}
    \centering
    \includegraphics[width=\linewidth]{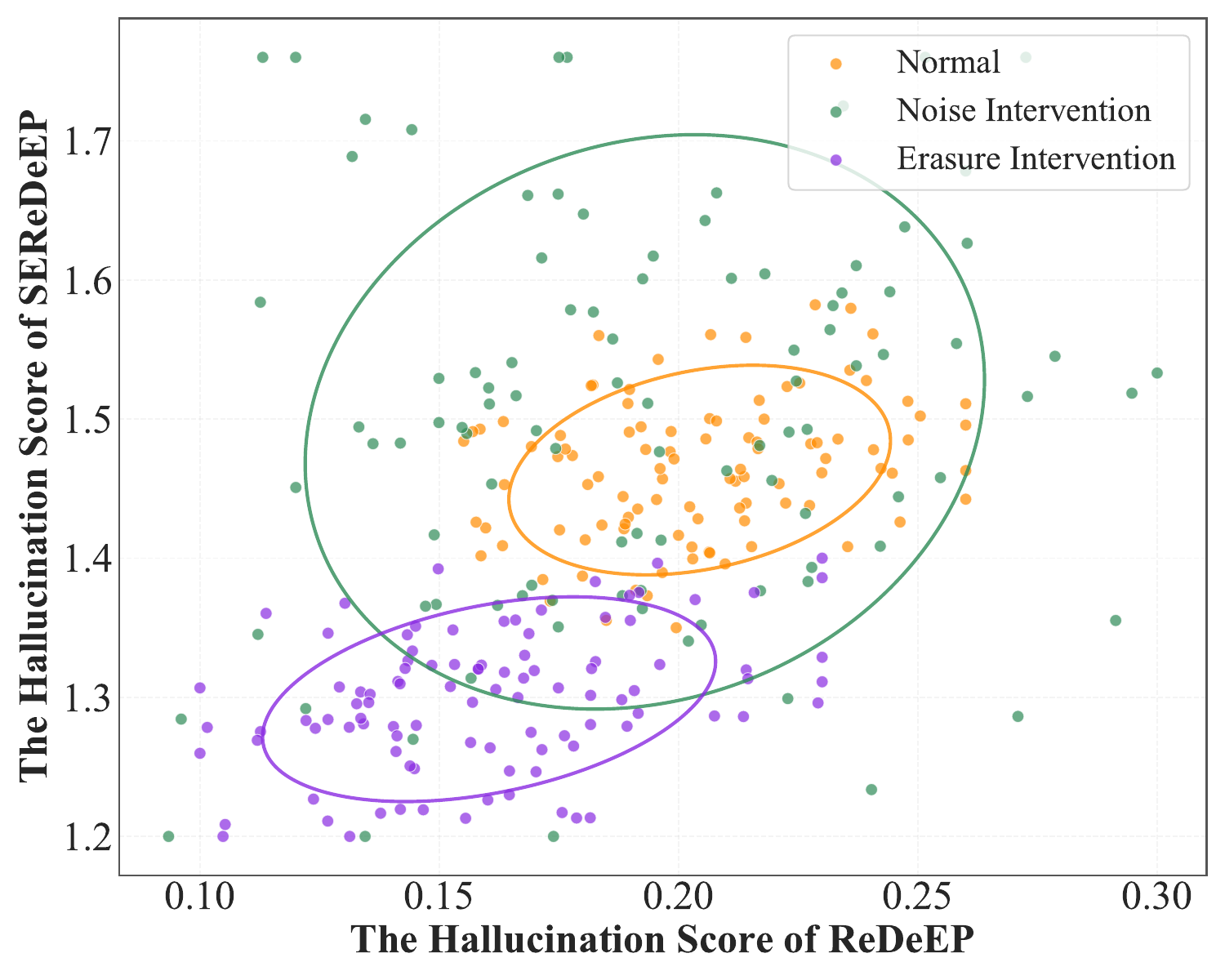}
    \caption{\centering{\textit{The result of Intervention Experiments}}}
    \label{fig:clusters}
\label{fig5}
\end{wrapfigure}

\textbf{Figure \ref{fig5}} shows the distribution of hallucination scores before and after interventions. Compared to the control group, the experimental group receiving noise intervention exhibited significant fluctuations in hallucination scores, while the experimental group receiving erasure intervention showed a decrease in hallucination scores. However, both SEReDeEP and ReDeEP demonstrated strong aggregation relationships before and after interventions, indicating that both can evaluate hallucinations by quantifying the model's utilization of external context and internal knowledge.

\section{Conclusion}
We propose SEReDeEP, a scheme that uses semantic entropy to predict hallucinations from the internal states of RAG models. We embed pre-trained semantic entropy probes into the hidden states of RAG models to evaluate the utilization of external context and internal knowledge in the residual flow of the RAG model decoder, resulting in regression-decoupled hallucination scores. SEReDeEP predicts hallucinations at the semantic level, with computational costs far lower than other sampling and retrieval-based methods, while ensuring the accuracy of prediction results.

\bibliography{references}

\appendix

\section{Technical Background of Retrieval-Augmented Model (RAG) Decoder}
\label{Sec:A}
The decoder of RAG models is part of the Transformer architecture. Pure decoder language models represented by GPT have achieved significant success\cite{grattafiori2024llama,gallifant2024peer}. This paper focuses on studying RAG models with pure decoder architectures, and observing their internal working modes also helps us better analyze RAG hallucinations.

The following introduces the technical background of the decoder structure. Note that the calculation process described here has been greatly simplified, serving only as a conceptual supplement to the main text by listing key computational processes. In reality, the internal structure and training methods of the model are much more complex\cite{explainability}.

For the current input statement $\boldsymbol{i}$, resulting in response $\boldsymbol{r}$, the decoder model can be simply denoted as:
\begin{equation}
    \boldsymbol{r}=f_L(\boldsymbol{i})
\end{equation}

At the input end, $\boldsymbol{i}$ is divided into an embedding sequence $(\boldsymbol{x}_1,\boldsymbol{x}_2,...,\boldsymbol{x}_n)$ by word units, where $n=|i|$ is the sequence length. Next, the embedding sequence, as the model's hidden values, is sent to $L$ hidden layers for decoding operations.

In each layer, the hidden state receives influences from attention, feed-forward networks, and itself, then updates its state. This information transmission process is also known as residual flow\cite{elhage2021mathematical,ferrando2024information,wu2024retrieval}, which introduces residual networks into the original Transformer architecture.

The contribution of attention to the residual flow is determined by the attention weight matrix\cite{elhage2021mathematical}. For the $h$-th head of the $l$-th layer, $\boldsymbol{W}_Q^{l,h}\in\mathbb{R}^{d \times d_h}$ and $\boldsymbol{W}_K^{l,h}\in\mathbb{R}^{d \times d_h}$ are the query $(Q)$ and key $(K)$ projection weight matrices\cite{elhage2021mathematical}, respectively, used to project input vectors. The attention weight matrix is calculated as:
\begin{equation}
    \boldsymbol{a}_{n}^{l,h} = \text{softmax} \left( \frac{(\boldsymbol{x}^{l-1}_n \boldsymbol{W}^{l,h}_Q)(\boldsymbol{X}^{l-1} _{n-1}\boldsymbol{W}^{l,h}_K)^T}{\sqrt{d_h / {H}}} \right)
    \label{eq:a}
\end{equation}

Between layers $l-1$ and $l$, $\boldsymbol{W}_V^{l,h}\in\mathbb{R}^{d \times d_h}$ and $\boldsymbol{W}_O^{l,h}\in\mathbb{R}^{d_h \times d}$ are the value projection matrix and output fusion matrix, respectively\cite{kobayashi2021incorporating}. The former receives data from the attention weight matrix and projects the input hidden values into output values, while the latter fuses the output values from each head to obtain the final hidden values. The attention contribution and residual flow update principle from layer $l-1$ to layer $l$ are:
\begin{gather}
    \text{Attn}^{l,h}\left(\boldsymbol{x}_{n}^{l-1}\right) = \sum_{j \leq n} \boldsymbol{a}_{n,j}^{l,h} \boldsymbol{x}_j^{l-1} \boldsymbol{W}_V^{l,h} \boldsymbol{W}_O^{l,h} \nonumber\\
    \boldsymbol{x}_n^{l,\text{attn}} = \text{LayerNorm}\left( \boldsymbol{x}_n^{l-1}+\text{Attn}^{l,h}(\boldsymbol{x}_{n}^{l-1})\right)
\end{gather}

This is the self-attention module combining multi-head attention mechanism and $OV$ circuits\cite{kobayashi2021incorporating}, which provides information about the current context's attention situation. The self-attention module is always followed by a feed-forward network module, which injects the model's internal knowledge into the hidden state:
\begin{gather}
    \text{FFN}^l(\boldsymbol{x}_n^{l,\text{attn}}) = \boldsymbol{W}_2^l \cdot \text{ReLU}\left( \boldsymbol{W}_1^l \boldsymbol{x}_n^{l,\text{attn}} + \boldsymbol{b}_1^l \right) + \boldsymbol{b}_2^l \nonumber\\
    \boldsymbol{x}_n^{l} =  \text{LayerNorm}\left(\boldsymbol{x}_n^{l,\text{attn}}+\text{FFN}^l(\boldsymbol{x}_n^{l,\text{attn}}) \right)\label{eq:ffn}
\end{gather}

where $\boldsymbol{x}_n^{\text{mid}, l} \in \mathbb{R}^d$ is the intermediate input vector of the $l$-th layer, coming from the output of the multi-head attention sublayer, $\boldsymbol{W}_1^l \in \mathbb{R}^{d_{\text{ff}} \times d}$ and $\boldsymbol{W}_2^l \in \mathbb{R}^{d \times d_{\text{ff}}}$ are internal parameter weight matrices used for linear transformation of hidden values. $\text{LayerNorm}()$ is the layer normalization operation, containing learnable parameters $\boldsymbol{\gamma} \in \mathbb{R}^d$ (scaling) and $\boldsymbol{\beta} \in \mathbb{R}^d$ (translation).

After the last hidden layer, activation values are mapped to probabilities in the vocabulary\cite{vaswani2017attention,zheng2025self}:
\begin{equation}
P(w_t | w_{1:t-1}) = \text{Softmax}(\boldsymbol{W}_o \,\boldsymbol{x}_n^L + \boldsymbol{b}_o)
\end{equation}

where $\boldsymbol{W}_o \in \mathbb{R}^{|V| \times L}$ and $\boldsymbol{b}_o \in \mathbb{R}^{|V|}$ are the output projection matrix and bias term, respectively, and $|V|$ is the vocabulary size.

Mainstream decoders use either greedy search or beam search to predict the final sequence\cite{boulanger2013audio,sutskever2014sequence,graves2012sequence}. In our experiments, we chose beam search. At each step, defining the beam width as $B$, i.e., selecting the $B$ tokens with the highest probability, maintaining a set of candidate sequences, the recursive formula is:
\begin{equation}
S_{t+1} = \underset{\substack{(y_{1:t}, s) \in S_t \\ w \in V}}{\text{Top-B}} \left[ s + \log P(w|y_{1:t}) \right]
\end{equation}

Select the sequence with the highest normalized score from the list:
\begin{equation}
r = \underset{y \in \mathcal{C}}{\text{argmax}} \left( \frac{1}{L_y^\alpha} \sum_{t=1}^{L_y} \log P(y_t|y_{1:t-1}) \right)
\label{r}
\end{equation}

where $\mathcal{C}$ is the set of completed sequences, $L_y$ is the sequence length, and $\alpha$ is the length penalty coefficient used to balance the length of sentences generated by the model.

\section{ Supplementary Information on ReDeEP's Working Principle\cite{ReDeEP}}
\label{Sec:ReDeEP}
This section supplements the implementation details of ReDeEP. Token-level ReDeEP has already been introduced in the main text; this method is simple in principle and relatively low in cost. Chunk-level ReDeEP is used to improve the token-level ReDeEP's lack of sufficient consideration for context\cite{fan2024survey,finardi2024chronicles}.

The inspiration for chunk-level ReDeEP comes from the common "chunking" strategy in RAG models, which divides the response $\boldsymbol{r}$ and context $\boldsymbol{c}$ into manageable segments, then performs average pooling on these blocks in the attention weight matrix to determine the highest-scoring attention blocks $(\boldsymbol{\widetilde{r}},\boldsymbol{\widetilde{c}})$. The chunk-level context score is calculated using the following method:

\begin{equation}
      \tilde{\mathcal{E}}_{\mathbf{r}}^{l} = \frac{1}{M} \sum_{\tilde{\mathbf{r}} \in \mathbf{r}} \tilde{\mathcal{E}}_{\tilde{\mathbf{r}}}^{l}, \quad \tilde{\mathcal{E}}_{\tilde{\mathbf{r}}}^{l} = \frac{\text{emb}(\tilde{\mathbf{r}}) \cdot \text{emb}(\tilde{\mathbf{c}})}{\|\text{emb}(\tilde{\mathbf{r}})\| \|\text{emb}(\tilde{\mathbf{c}})\|} 
\end{equation}

For chunk-level internal parametric knowledge, token-level parametric knowledge is calculated for each chunk and then summed:

\begin{equation}
\tilde{\mathcal{P}}_{\mathbf{r}}^{l} = \frac{1}{M} \sum_{\tilde{\mathbf{r}} \in \mathbf{r}} \tilde{\mathcal{P}}_{\tilde{\mathbf{r}}}^{l}, \quad \tilde{\mathcal{P}}_{\tilde{\mathbf{r}}}^{l} = \frac{1}{|\tilde{\mathbf{r}}|} \sum_{t \in \tilde{\mathbf{r}}} \mathcal{P}_{t}^{l}
\end{equation}

When calculating the hallucination score, linear regression is still used to combine the two covariates. Based on this, some operations are performed to mitigate RAG hallucinations. A hallucination threshold $\tau$ is selected, and when the hallucination score of the current token exceeds $\tau$, the weights of the attention heads and FFN modules are adjusted, replacing $\text{Attn}^{l,h}\left(x_{n}^{l-1}\right)$ with $\mu \text{Attn}^{l,h}\left(x_{n}^{l-1}\right)$, and replacing $\text{FFN}^l(\boldsymbol{x}_n^{l,\text{attn}})$ with $\nu \text{FFN}^l(\boldsymbol{x}_n^{l,\text{attn}})$. Here, $\mu$ is a constant greater than 1, used to amplify the contribution of attention heads, and $\nu$ is a constant between 0 and 1, used to suppress the contribution of FFN.

\section{RAG Amplifies the Instability of Non-Semantic Entropy Detection Methods After Introducing External Knowledge}
This section discusses how the influx of external knowledge in RAG amplifies the bias of non-semantic entropy detection methods.

Similarly, we use the datasets mentioned in the main text to conduct experiments on LLaMA2-7B\cite{metallama2023} and LLaMA3-7B\cite{metallama2024}. In addition to using semantic entropy probes to evaluate hallucinations, we also employ methods such as TLE (naive token-level entropy), FC (factual consistency)\cite{kryscinski2019evaluating}, P(true) (credibility estimation of generated answers)\cite{kadavath2022language}, AP (accuracy probe), and LLH (log-likelihood). We design two groups: one group enables RAG technology(such as ChromaDB\cite{chromadb_pypi}) during the generation process to enhance semantic understanding, while the other group blocks knowledge acquisition functionality, relying solely on the model's internal original parameters to answer. We then calculate the accuracy rates for each experimental group.

The results in \textbf{Table \ref{tab:comparison}} show that, compared to using semantic entropy measures, other measurement schemes exhibit larger discrepancies in accuracy between models with RAG enabled and disabled, and these discrepancies demonstrate irregularity. This may be because the interference of externally retrieved knowledge changes the semantic expression of model responses, although their semantics are actually similar. The SE metric did not show significant fluctuations, which further validates the rationality of introducing semantic entropy to evaluate model hallucinations.

\begin{table}[!ht]
\centering
\caption{\textit{Comparison before and after enabling RAG}}
\label{tab:comparison}
\begin{tabular}{cc cccccc}  
\toprule
\multirow{2.5}{*}{\textbf{Model}} & \multirow{2.5}{*}{\textbf{STATUS}} & \multicolumn{6}{c}{\textbf{Method ACC}} \\ 
\cmidrule(lr){3-8}
& & \textbf{TLE} & \textbf{FC} & \textbf{P(true)} & \textbf{AP} & \textbf{LLH} & \textbf{SE} \\
\midrule
\multirow{2}{*}{\textbf{LLaMa2-7B}\cite{metallama2023}} & \textit{OFF} & 0.6942 & 0.6218 & 0.6766 & 0.7234 & 0.6527 & 0.7129 \\
                        & \textit{ON} & 0.7857 & 0.7506 & 0.7019 & 0.6851 & 0.7003 & 0.7193 \\
\addlinespace
\multirow{2}{*}{\textbf{LLaMa3-7B}\cite{metallama2024}} & \textit{OFF} & 0.6884 & 0.7592 & 0.7017 & 0.6684 & 0.7316 & 0.7257 \\    
                        & \textit{ON} & 0.6135 & 0.6587 & 0.6782 & 0.6401 & 0.7099 & 0.7301 \\    
\bottomrule
\end{tabular}
\end{table}

\section{Training Process of Semantic Entropy Probes\cite{SEP}}
\label{Sec:SEP}
\textbf{Semantic Entropy Acquisition.} We used the Prompt shown below to DeepSeek-R1 and Claude 3.7 Sonnet, expecting them to help us complete qualified semantic entropy predictions. When both DeepSeek-R1 and Claude 3.7 Sonnet provide bidirectional entailment results, we consider the two answers to have a bidirectional entailment relationship. These prediction data would be used to train our semantic entropy probes.
\newtcolorbox{promptbox}[1][]{
  colback=gray!10,
  colframe=gray!50,
  boxrule=0.5pt,
  arc=2pt,
  width=\linewidth, 
  enhanced, 
  boxed title style={colframe=gray!50,colback=gray!10}, 
  attach boxed title to top left={xshift=5mm,yshift=-2mm}, 
  #1
}
\begin{promptbox}
\textit{We have a question:\\\{Question\}\\Assuming the large laguage model generated the following two answers:\\\{Answer \text{1}\}\\\{Answer \text{2}\}\\Determine the entailment relationship between Answer \text{1} and Answer \text{2}.\\If Answer \text{1} entails Answer \text{2}, output \text{2};\\If Answer \text{2} entails Answer \text{1}, output \text{1};\\If they entail each other (bidirectional entailment), output \text{0};\\If they are unrelated, output \text{3}.} 
\label{D}
\end{promptbox}

\textbf{Probe Training.} As mentioned earlier, the semantic entropy probe is a logistic regression classifier placed on the Transformer's hidden layers. We need to train it using pre-known semantic entropy. The semantic entropy probe is a linear probe.

For the current hidden state $\boldsymbol{x}_n^l$, after processing through the classifier $g(\boldsymbol{x}_n^l)$, the logistic regression classifier is:
\begin{equation}
    \hat{H}_{\text{SE}}(x) = \mathbbm{1} \,[H_{\text{SE}}(x) > \gamma^{\star}]
\end{equation}

The determination of the threshold ${\gamma}^*$ uses a clustering idea similar to one-dimensional k-means. We calculate semantic entropy (discrete entropy) for output responses and classify semantic entropy as high or low using a preset threshold $\gamma$. By optimizing the objective function, we minimize the sum of squared errors between each sample's score in both groups and their corresponding means. The probe outputs the probability of being predicted as high, which can approximate semantic entropy.

\section{Supplementary Information on Experimental Background}
\label{sec:Baseline}
We selected nine methods, including SEP and ReDeEP, for our experiments. Below is a brief introduction to some of these methods:
\begin{itemize}
\item \textbf{AGSER\cite{AGSR}.} Utilizes attention mechanisms to guide large models in so-called "self-reflection."

\item \textbf{Ragas\cite{RAGAS}.} Breaks down responses into shorter clauses and calculates a faithfulness score for each clause.

\item \textbf{ITI\cite{ITI}.} Employs a binary classifier to "score" the model's attention heads and response performance.

\item \textbf{InterrogateLLM\cite{yehuda2024interrogatellm}.} Detects hallucinations by sending reverse questions to the LLM and verifying the consistency of generated content with the original query.

\item \textbf{INSIDE\cite{inside}.} By mining the internal state information during the generation process of LLMs, combined with EigenScore and feature pruning at test time, efficient hallucination detection was achieved.

\item \textbf{FActScore\cite{min2023factscore}.} Through fine-grained retrieval and verification, decomposes generated long texts into atomic facts and calculates the proportion of each fact supported in external knowledge bases.

\item \textbf{SelfCheckGPT\cite{selfcheckgpt}.} Makes the model generate multiple responses and measures consistency between responses to identify fictional content, determining inconsistent ones as hallucinations.
\end{itemize}

\section{Cost and Overhead Analysis}
Under conditions that ensure its reliable performance, the semantic entropy probe is very cost-effective because it is simply an unsupervised classification method. After obtaining training data for the semantic entropy probe, training can be completed in a short time with just a CPU, and its performance reaches the level of many complex models.

Because the probe is directly embedded into the hidden layers of the language model decoder, the calculation speed of the SEReDeEP hallucination score is also very fast, continuing the advantage of ReDeEP in terms of time overhead.

\section{More Ablation}
The ablation experiments mentioned in the main text \textbf{ Section \ref{ablation}}. It verified the comprehensive improvement of model performance after introducing semantic entropy, both in terms of external context and parametric knowledge.
\label{More}
\begin{table}[htbp]
\centering
\fontfamily{ptm}\selectfont
\caption{\textit{Ablation of SEReDeEP and ReDeEP}}
\label{tab:more ablation}
\footnotesize 
\setlength{\tabcolsep}{3pt} 
\renewcommand{\arraystretch}{1.1}
\begin{tabular}{@{} 
    lc 
    *{4}{c} 
    *{4}{c} 
@{}}
\toprule[0.8pt]
\multirow{2.5}{*}{\textbf{Model}} & 
\multirow{2.5}{*}{\textbf{CONDITION}} & 
\multicolumn{4}{c}{\textbf{RAGTruth}\cite{ragtruth}} & 
\multicolumn{4}{c}{\textbf{HalluRAG}\cite{ridder2024hallurag}} \\
\cmidrule(lr){3-6} \cmidrule(l){7-10}
&& 
\textbf{ACC} & \textbf{AUC} & \textbf{F1} & \textbf{Rec} & \textbf{ACC} & \textbf{AUC} & \textbf{F1} & \textbf{Rec} \\ 
\midrule[0.6pt]

\multirow{5}{*}{\centering \textbf{LLaMA3-7B}\cite{metallama2024}}
& ECS \ref{ECS} & 0.8312 & 0.6846 & 0.6532 & 0.6931 & 0.7855 & 0.6273 & 0.6591 & 0.6908 \\
& PKS \ref{PKS} & 0.8269 & 0.7217 & 0.7045 & 0.6716 & 0.8006 & 0.7042 & 0.7013 & 0.7082 \\
& ECS+PKS \ref{ECS} \ref{PKS} & 0.8364 & 0.7269 & 0.7124 & 0.6938 & 0.8026 & 0.7156 & 0.7094 & 0.7102 \\
& ECS+\textbf{PKE} \ref{ECS} \ref{PKE} & 0.8456 & 0.7435 & 0.7169 & 0.7028 & 0.8145 & 0.7437 & 0.7291 & 0.7153 \\
& \textbf{ECE}+PKS \ref{ECE} \ref{PKS} & 0.8401 & 0.7653 & 0.7199 & 0.7064 & 0.8057 & 0.7415 & 0.7123 & 0.7128 \\
\specialrule{0.4pt}{4pt}{0pt} 

\multirow{5}{*}{\centering \textbf{Mistral-7B}\cite{mistral2024}}
& ECS \ref{ECS} & 0.7823 & 0.6179 & 0.6484 & 0.6729 & 0.7214 & 0.6836 & 0.6303 & 0.6815 \\
& PKS \ref{PKS} & 0.7694 & 0.6285 & 0.6457 & 0.6612 & 0.7531 & 0.6412 & 
0.6123 & 0.6938 \\
& ECS+PKS \ref{ECS} \ref{PKS} & 0.8012 & 0.6527 & 0.6633 & 0.6894 & 0.7618 & 0.6925 & 0.6376 & 0.7106 \\
& ECS+\textbf{PKE} \ref{ECS} \ref{PKE} & 0.8321 & 0.7039 & 0.6894 & 0.6906 & 0.7667 & 0.7429 & 0.7038 & 0.7143 \\
& \textbf{ECE}+PKS \ref{ECE} \ref{PKS} & 0.8319 & 0.7427 & 0.7034 & 0.6792 & 0.7613 & 0.7233 & 0.6891 & 0.7154 \\
\specialrule{0.4pt}{4pt}{0pt} 

\multirow{5}{*}{\centering \textbf{Qwen2.5-7B}\cite{Qwen2024}}
& ECS \ref{ECS} & 0.6253 & 0.7328 & 0.6201 & 0.6125 & 0.7839 & 0.5527 & 0.6281 & 0.6854 \\
& PKS \ref{PKS} & 0.7316 & 0.7003 & 0.6219 & 0.6612 & 0.7136 & 0.6539 & 0.6722 & 0.7084 \\
& ECS+PKS \ref{ECS} \ref{PKS} & 0.7592 & 0.7520 & 0.6552 & 0.6847 & 0.7980 & 0.6631 & 0.6839 & 0.7091 \\
& ECS+\textbf{PKE} \ref{ECS} \ref{PKE} & 0.8023 & 0.7647 & 0.6953 & 0.6992 & 0.8237 & 0.7439 & 0.7035 & 0.7124 \\
& \textbf{ECE}+PKS \ref{ECE} \ref{PKS} & 0.8234 & 0.7453 & 0.6736 & 0.6836 & 0.8396 & 0.7012 & 0.7294 & 0.7103 \\

\bottomrule[0.8pt]
\end{tabular}
\end{table}

\end{document}